\pdfoutput=1

\documentclass[10pt,twocolumn,letterpaper]{article}

\usepackage{cvpr}              
\usepackage{graphicx}
\usepackage{amsmath}
\usepackage{amssymb}
\usepackage{booktabs}
\usepackage{multirow}
\usepackage{gensymb} 
\usepackage[outdir=./]{epstopdf}
\usepackage[toc,page]{appendix}
\usepackage[pagebackref,breaklinks,colorlinks]{hyperref}

\usepackage[capitalize]{cleveref}
\crefname{section}{Sec.}{Secs.}
\Crefname{section}{Section}{Sections}
\Crefname{table}{Table}{Tables}
\crefname{table}{Tab.}{Tabs.}

\def\cvprPaperID{18} 
\def\confName{CVPR}
\def\confYear{2023}

\newcommand\blfootnote[1]{%
  \begingroup
  \renewcommand\thefootnote{}\footnote{#1}%
  \addtocounter{footnote}{-1}%
  \endgroup
}

\begin{document}

\title{OCTraN: 3D Occupancy Convolutional Transformer Network in Unstructured Traffic Scenarios}

\author{
Aditya Nalgunda Ganesh \and Dhruval Pobbathi Badrinath \and Harshith Mohan Kumar \and Priya SS \qquad Surabhi Narayan\\ 
Department of Computer Science, PES University, Bengaluru\\
{\tt\small \{adityang5, dhruvalpb, hiharshith18, sspriya147\}@gmail.com surabhinarayan@pes.edu}
}

\maketitle
\blfootnote{This work was accepted as a spotlight presentation at the \href{https://sites.google.com/view/t4v-cvpr23/papers}{Transformers for Vision Workshop @CVPR 2023}.\\}
\begin{abstract}
Modern approaches for vision-centric environment perception for autonomous navigation make extensive use of self-supervised monocular depth estimation algorithms that output disparity maps. However, when this disparity map is projected onto 3D space, the errors in disparity are magnified, resulting in a depth estimation error that increases quadratically as the distance from the camera increases. Though Light Detection and Ranging (LiDAR) can solve this issue, it is expensive and not feasible for many applications. To address the challenge of accurate ranging with low-cost sensors, we propose---OCTraN---a transformer architecture that uses iterative-attention to convert 2D image features into 3D occupancy features and makes use of convolution and transpose convolution to efficiently operate on spatial information. We also develop a self-supervised training pipeline to generalize the model to any scene by eliminating the need for LiDAR ground truth by substituting it with pseudo-ground truth labels obtained from boosted monocular depth estimation. Our code\footnote{Code: \url{https://bit.ly/OCTraN-Code}} and dataset\footnote{Dataset: \url{https://bit.ly/OCTraN-Dataset}} have been made public.
\end{abstract}
\section{Introduction}
\label{sec:intro}
Autonomous navigation relies on dense 3D scene reconstruction for action in physical environments \cite{doi:10.1126/scirobotics.aaw6661}. Recently, self-attention based architectures inspired by \cite{ViT} have begun outperforming convolutional neural networks in 3D object detection, semantic scene segmentation, monocular depth prediction, and other current vision perception tasks. Particular advances in transformers demonstrate that these models \cite{brown2020language_gpt3, bert, jaegle2021perceiver, zhao2021point, ViT} are excellent domain agnostic learners at the expense of having no good inductive priors for any given domain.

A caveat of image based ranging techniques that use transformers is that they require large volumes of LiDAR depth data synchronized with corresponding RGB images. While accurate, LiDAR units are expensive and produce sparse maps of the environment. To tackle the data scale problem, we introduce a depth dataset generation pipeline that eliminates the need for expensive ranging sensors by substituting them with depth labels obtained from boosted monocular depth estimation \cite{Miangoleh2021Boosting}. Existing datasets with depth data from LiDAR sensors \cite{KITTI, nuScenes, mao2021one, packnet, DBLP:journals/corr/abs-2109-13410, doi:10.1177/0278364920961451, pitropov2021canadian,geyer2020a2d2,astar-3d, RadarRobotCarDatasetICRA2020, RobotCarDatasetIJRR, Ettinger_2021_ICCV}, are primarily focused on driving environments with predictable and structured traffic. The lack of variety involving complex dense unstructured traffic introduces bias resulting in a data distribution shift toward structured environments. Therefore current state-of-the-art models \cite{ranftl2020robust_midas, ranftl2021vision_DPT, yin2020learning_leres, manydepth, monodepth2} fail to generalize in developing countries where traffic flow is unpredictable. To mitigate this issue we have gathered a small dataset in Bengaluru---with elements of complex unstructured traffic---using our proposed pipeline.

With the goal of improving the robustness and generalization of vision based depth perception models, we propose OCTraN. This transformer network inherits features extracted from a convolution-based backbone, providing good inductive priors for (2D and 3D) spatial information. It then uses iterative-attention as a feature-space transformer to transform 2D image features into 3D occupancy features. Finally, this feature space is transformed by a convolution transpose block to produce 3D occupancy.


\begin{figure*}[th!]
\centering
   \includegraphics[width=\linewidth]{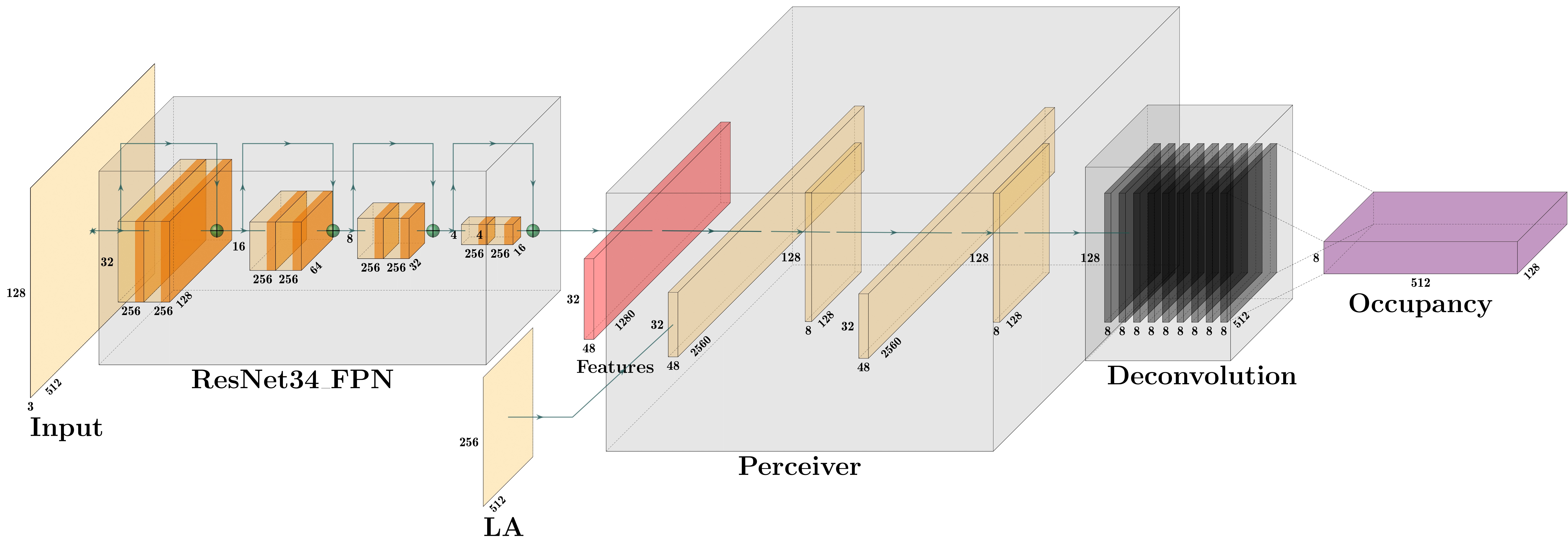}
  \caption{Architectural diagram of the 3D Occupancy Convolutional Transformer Network (OCTraN).}
  \label{fig:modelArchitecture}
\end{figure*}
\section{Related Work}
\label{sec:relate}
\textbf{Self-Supervised Depth Perception.} Owing to the difficulty in obtaining ground truth depth for varied real-world environments, several self-supervised depth perception algorithms have been proposed \cite{monodepth, monodepth2, manydepth, megadepth, gcndepth}. While such models require only a single camera, making real-world deployment easy, they have a multitude of issues yet to be addressed. They produce disparity maps that are locally and temporally inconsistent. Watson \etal \cite{manydepth} overcame the temporal inconsistency by utilizing multiple consecutive frames as input. However, since disparity is inversely proportional to depth as shown in \cref{eqTriang}, for far-away objects that have a lower disparity value, minute variations in disparity correspond to large variations in depth, resulting in non-uniform resolution point clouds with closer objects being represented with more points compared to those further away. This makes safe path planning a challenge.

\textbf{Bird's Eye View (BEV) Architectures.} A top-down view of a scene gives a global perspective of the surrounding environment and captures static and dynamic elements well. BEV architectures like \cite{reiher2020sim2real, li2022bevformer, roddick2020predicting} provide this top-down map that can be used to perform path planning. The idea of predicting BEV from multiple camera views has demonstrated similar performance to LiDAR-centric methods \cite{zhu2020cylindrical, liu2022bevfusion}. The limitation of this approach is that it fails to account for 3D information about the scene such as unclassified objects, potholes, and overhanging obstacles. 

\textbf{3D Occupancy Networks.} 
Obtaining an effective representation for a 3D scene is a fundamental task in 3D environment perception. One direct way is to discretize the 3D space into voxels in an occupancy grid \cite{zhou2017voxelnet,zhu2020cylindrical}. The ability to describe fine-grained 3D structures makes voxel-based representation favorable for 3D semantic occupancy prediction tasks including LiDAR segmentation \cite{cheng2021af2s3net, liong2020amvnet, tang2020searching, ye2022lidarmultinet, ye2021drinet, zhu2020cylindrical} and 3D scene completion \cite{cao2022monoscene, chen20203d, li2020anisotropic, roldão2020lmscnet, yan2020sparse}. A recent approach, TPVFormer \cite{huang2023triperspective} optimizes the memory by representing the 3D space as projections on 3 orthogonal planes. While these approaches have made admirable strides, they don't aim to address the bias of existing datasets toward structured traffic. 

\section{Proposed Work}
\begin{figure}
    \centering
    \begingroup
    \newcommand*\rot{\rotatebox{90}}
    \setlength{\tabcolsep}{1pt} 
    \renewcommand{\arraystretch}{0.5} 

  \begin{tabular}{lll}
\rot{\hspace{4.5mm}{\scriptsize RGB 
\textcolor{white}{[]} 
}}
\includegraphics[width=.30\columnwidth]{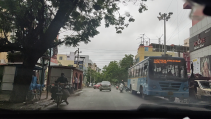} & 
\includegraphics[width=.30\columnwidth]{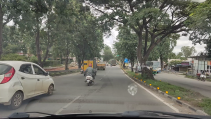} & 
\includegraphics[width=.30\columnwidth]{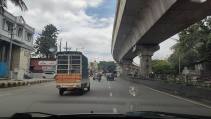}\\

\rot{\hspace{0.5mm}{\scriptsize MiDaS
\cite{ranftl2020robust_midas}
}}
\includegraphics[width=.30\columnwidth]{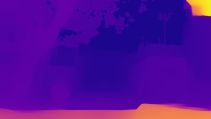} & 
\includegraphics[width=.30\columnwidth]{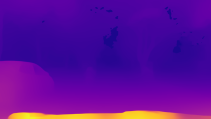} & 
\includegraphics[width=.30\columnwidth]{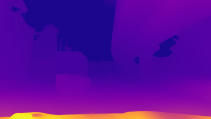}\\

\rot{\hspace{0mm}{\scriptsize Boosted
\cite{Miangoleh2021Boosting}
}}
\includegraphics[width=.30\columnwidth]{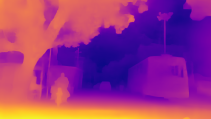} & 
\includegraphics[width=.30\columnwidth]{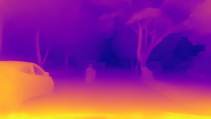} & 
\includegraphics[width=.30\columnwidth]{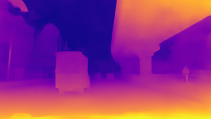}\\

  \end{tabular}
  \endgroup
  \caption{In these samples from our dataset, we illustrate the difference between disparity maps produced by monocular depth estimation MiDaS v3.0 $DPT_{L-384}$ \cite{ranftl2020robust_midas, Ranftl2021_DPT} and the boosting technique \cite{Miangoleh2021Boosting}. Boosting can extract depth information of far-off objects and fine details of closer objects.}
  \label{fig:dataset_samples} 
\end{figure}
\subsection{Depth Dataset for Unstructured Traffic}
\begin{figure}
    \begingroup
    \setlength{\tabcolsep}{0pt} 
    \renewcommand{\arraystretch}{0.5} 
    \begin{tabular}{ll}
    \includegraphics[width=.41\linewidth]{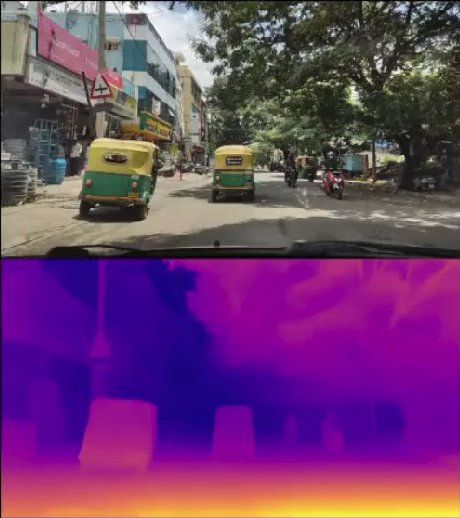}& 
    \includegraphics[width=.58\linewidth]{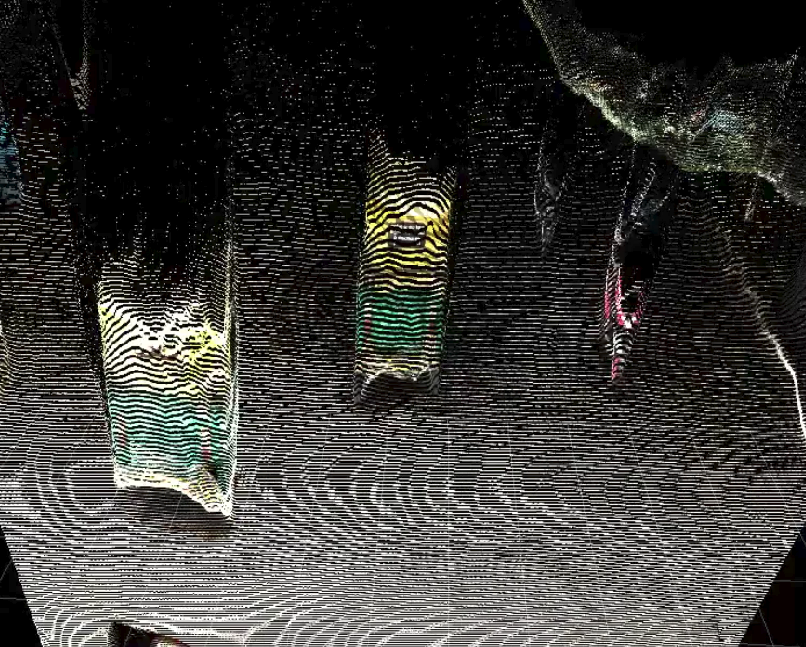}\\
    \end{tabular}
    \endgroup
    \caption{On the left is an RGB frame and corresponding disparity map from our dataset. On the right is the disparity map being projected into 3D space using the camera calibration. The two rickshaws and the bike are clearly visible in 3D space}
    \label{fig:point_cloud} 
\end{figure}

We gathered a dataset spanning 114 minutes and 165K frames in Bengaluru, India. Our dataset consists of video data from a calibrated camera sensor with a resolution of 1920×1080 recorded at a framerate of 30 Hz. We utilize a Depth Dataset Generation pipeline that only uses videos as input to produce high-resolution disparity maps, as shown in \cref{fig:dataset_samples}, by boosting monocular disparity estimation systems \cite{Miangoleh2021Boosting}. The Appendix offers more details about this depth generation. We project the produced disparity maps into 3D space using the camera intrinsics as shown in \cref{eqTriang} where $(x,y,z)$ are the coordinates of the point in the camera coordinate frame, ($u$, $v$) are the pixel coordinates in the image, ($f_x$, $f_y$) are the effective focal lengths of the camera in pixels along the $x$ and $y$ directions respectively and $(o_x, o_y)$ is the sensor's principle point. The function $d(u,v)$ is the disparity value at the pixel coordinates $(u,v)$. The term $b$ is the virtual baseline of the sensor setup. \cref{fig:point_cloud} shows a sample 3D projection from an image in our dataset.

\label{sec:proposed}
\subsection{OCTraN Architecture}
As shown by Jaegle \etal \cite{jaegle2021perceiver}, the Attention mechanism is an excellent domain agnostic learner with the drawback of scaling quadratically with input size $(M)$. The complexity of the Queries-Keys-Values (QKV) attention operation \cite{vaswani2017attention} as shown in \cref{attention_eqn} – is $O(M^2)$. Therefore, it is infeasible to apply attention to the entire image at once.
\begin{equation}\label{attention_eqn}
Attention(Q, K, V) = softmax(\frac{QK^T}{\sqrt{d_k}})V
\end{equation}
Our architecture as shown in \cref{fig:modelArchitecture} is inspired by the Perceiver \cite{jaegle2021perceiver}. We set the queries $Q$ as a projection of a learned latent array with index dimension $N<<M$, resulting in a cross-attention operation of complexity $O(MN)$. To upsample the occupancy grid produced by the transformer, we have a series of Convolution Transpose layers that have good inductive priors for operating on 3D spatial data.

In OCTraN-B, the transformer is tasked with extracting 2D image features and converting them into 3D occupancy features. In OCTraN-V0, we leverage an image feature extraction backbone consisting of ResNet \cite{he2015deep} in combination with a Feature Pyramid Network (FPN) \cite{lin2017feature}. The convolutional backbone is able to efficiently extract image features from the 2D spatial data leaving the transformer with the sole task of 2D to 3D feature conversion. The OCTraN-V1 architecture further associates each column in 2D image space with a cuboid in 3D occupancy space, allowing the transformer to attend to smaller segments of the task. This is similar to \cite{roddick2020predicting} where each column in image space corresponds to a ray in BEV. To extract the chunked image features $F_i^j$ for chunk $i$ out of $C$ total chunks, we apply \cref{chunking_eqn}. Where $P$ is the feature pyramid whose layers have shapes defined by $\{P^{j} \in \mathbf{R}^{2^{j+1} \times 2^{j+3} \times 256} \forall j \in [0,4] \}$ and $j$ is the layer number of the pyramid. The feature shapes of $P$ are shown in \cref{fig:modelArchitecture} as well. All the layers of the feature pyramid have 256 channels and their dimensions depend on the layer number $j$ and input shape ($128\times512$).
\begin{equation}
    \begin{split}
    \label{chunking_eqn}
    \centering
    F_i^j = P_{ab}^j \forall a \in [0,2^{j+1}), b \in [\frac{i}{C} 2^{j+3}, \frac{i+1}{C} 2^{j+3})
    \end{split}
\end{equation}

\begin{figure}
    \begingroup
    \setlength{\tabcolsep}{0pt} 
    \renewcommand{\arraystretch}{0.5} 
    \begin{center}
    \begin{tabular}{l}
    \includegraphics[width=\columnwidth]{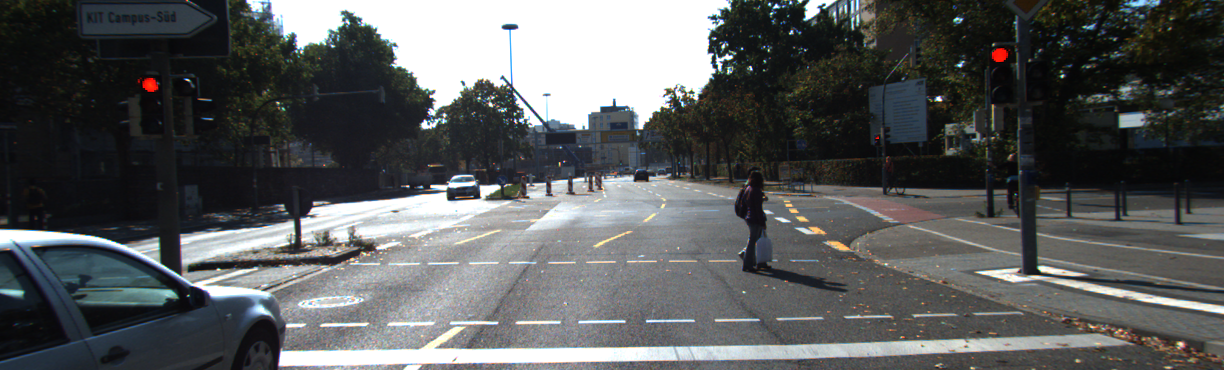}\\
    \includegraphics[width=\columnwidth]{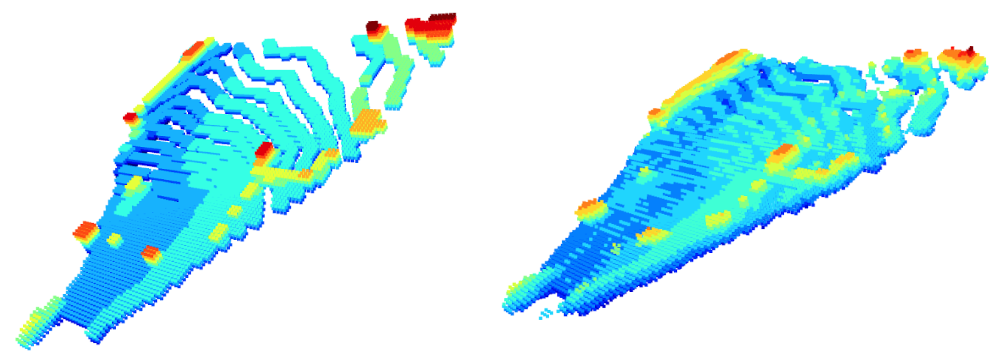}\\
    \end{tabular}
    \end{center}
    \endgroup
    \caption{An input image with corresponding ground truth occupancy grid obtained from LiDAR (left) and OCTraN prediction (right). The grid consists of $(128,128,8)$ voxels and covers $(85,51,4)$ meters and is colored by height.}
    \label{fig:model_ouputs} 
\end{figure}
As shown in \cref{fig:model_ouputs}, our system predicts in 3D occupancy space, unlike many existing monocular ranging techniques that predict in disparity space. Disparity is inversely proportional to depth as shown in \cref{eqDepthtoDisp}. This means that at a point on an image $(u,v)$ captured by a camera with a focal length $f_x$, a virtual baseline of $b$ and a disparity error $\Delta d$, the error in estimated depth $\Delta z$ increases quadratically with depth $z$. By predicting directly in depth space, we are reducing this systemic inefficiency. 
\begin{equation} \label{eqDepthtoDisp}
\begin{split}
z = \frac{b f_x}{d} \implies
\Delta z = \frac{z^{2}}{b f_x} \Delta d
\end{split}
\end{equation}

\begin{table*}[t]
    \centering
    \begin{tabular}{c l c c c c c c c c c c}
        \toprule
        \multirow{2}{*}{\bfseries Method} & 
        \multirow{2}{*}{\bfseries Dataset} &
        \multicolumn{9}{c}{\bfseries Hyperparameters} &
        \multirow{2}{*}{\bfseries IoU (\%)}
        \\ \cmidrule(lr){3-11}
            && CH&CDH&LMP&LR&BS&MF&D&LH&LDH
        \\ \cmidrule(lr){1-12}
        OCTraN-B & KITTI & 1 & 64 & 0 & 0.001 & 1 & 1000 & 1 & 8 & 32 & 18.818\\
        OCTraN-V0 & KITTI & 8 & 32 & 0.5 & 0.0001 & 2 & 500000 & 4 & 8 & 32 & \textbf{34.669}\\
        OCTraN-V1 & KITTI & 8 & 32 & 0.5 & 0.0001 & 4 & 500000 & 8 & 4 & 32 & 28.408\\
        \bottomrule
    \end{tabular}
    \caption{\textbf{Ablation Study} on our proposed architecture comparing the optimal hyperparameters and the IoU scores achieved.}
  \label{tab:modelComparision}
\end{table*}

\begin{equation} \label{eqTriang}
\begin{split}
(x,y,z) = (\frac{b(u-o_x)}{d(u,v)}, \frac{b f_x (v-o_y)}{f_y d(u,v)}, \frac{b f_x}{d(u,v)})
\end{split}
\end{equation}

\begin{table}[]
    \centering
    \begin{tabular}{c|c|c}
        \toprule
        Model & Dataset & IoU (\%)\\
        \cmidrule(lr){1-3}
        *TPVFormer \cite{huang2023triperspective} &  SemanticKITTI & 34.25\\
        *MonoScene \cite{cao2022monoscene} &  SemanticKITTI & 34.16\\
        *JS3C-Net \cite{yan2021sparse} &  SemanticKITTI  & 34.00\\
        *AICNet \cite{li2020anisotropic} &  SemanticKITTI & 23.93\\
        *3DSketch \cite{chen20203d} & SemanticKITTI  & 26.85\\
        *LMSCNet \cite{roldão2020lmscnet} &  SemanticKITTI & 31.38\\
        OCTraN-V0 &  KITTI & \textbf{34.669}\\
        OCTraN-V1 & KITTI &  28.408\\
        \bottomrule
    \end{tabular}
    \caption{The performance numbers of * models are obtained from Huang \etal \cite{huang2023triperspective} who evaluated the same on SemanticKITTI. These models are compared with OCTraN models trained on KITTI.}
    \label{tab:other_methods}
\end{table}
\section{Experiments}
\begin{figure}%
    \centering
    \subfloat{{\includegraphics[trim={0.7cm 0 0 0 0},width=0.464\columnwidth]{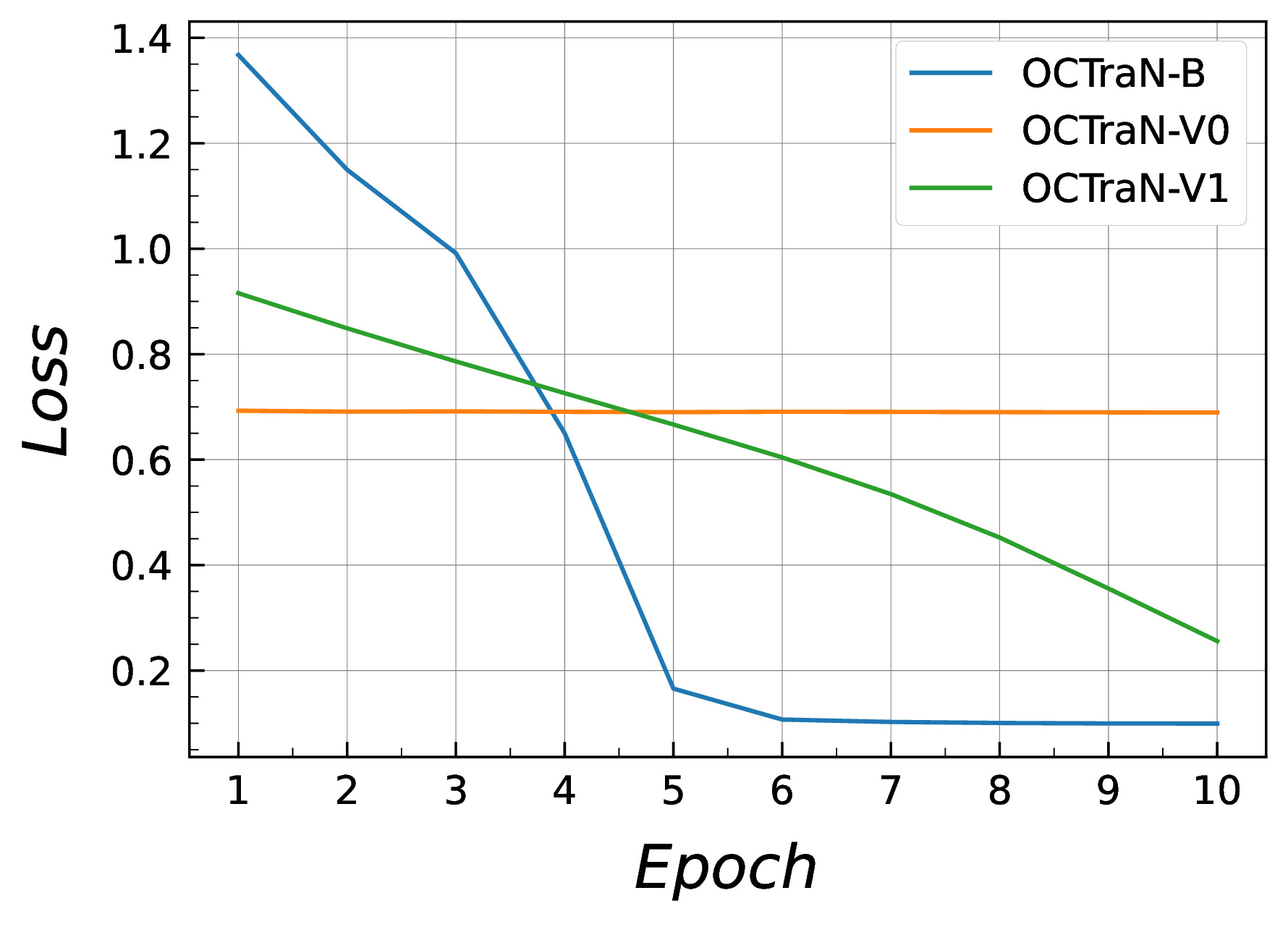} }}%
    \qquad
    \subfloat{{\includegraphics[trim={2cm 0 0 0 0},width=0.42\columnwidth]{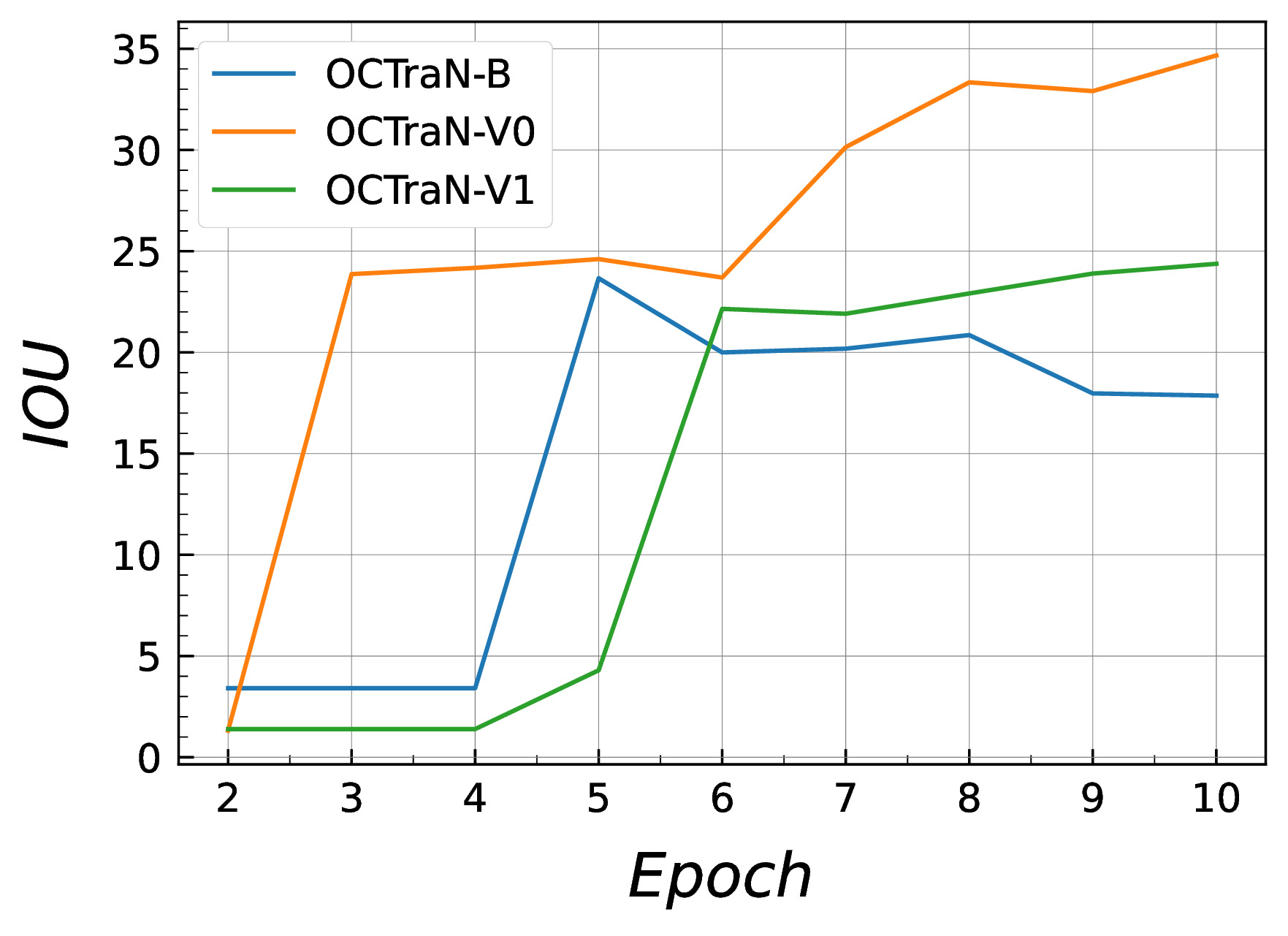} }}%
    \caption{Loss and IoU of the three architectures proposed.}%
    \label{fig:graphicalResults}%
\end{figure}
\label{sec:experiments}
\subsection{Experimental Setup}
Our experiments have been performed on the KITTI dataset \cite{geiger2012cvpr_kitti} using PyTorch 1.13.0 \cite{NEURIPS2019_9015_pytorch}. Our networks have been trained on an Nvidia RTX A6000. We utilized Weights \& Biases \cite{wandb} to track our experiments across multiple sweeps. In \cref{tab:modelComparision}, we compare multiple variants of our architecture, with optimal hyperparameters: cross heads (CH), cross-dimensional heads (CDH), loss mask probability (LMP), learning rate (LR), batch size (BS), max frequency (MF), depth (D), latent heads (LH) and latent dimensional heads (LDH). Performing experiments across multiple sweeps allowed for the identification of the most important hyperparameters and assisted in fine-tuning our architectures to achieve maximal performance. The Appendix offers more details about the hyperparameter sweep. 
In \cref{tab:modelComparision}, we report the most optimal hyperparameters and Intersection over Union (IoU) scores. The results from OCTraN-V0 and OCTraN-V1 indicate that extracting features using a ResNet backbone significantly boosts performance when compared to feeding raw images to the Perceiver. OCTraN-V0 performed the best amongst all three variations as it achieved a final IoU of 34.67\% after ten epochs. However, observing the trend of the loss over epochs in \cref{fig:graphicalResults} indicates the contrary. OCTraN-V0's loss decreases much slowly compared to OCTraN-V1's drastic decrease in loss indicating that chunking allows for better learning and with increased training time, OCTraN-V1 will converge with an optimal learned representation.
\subsection{Comparison with Existing Methods}
OCTraN-V0 is comparable to current state-of-the-art networks as shown in \cref{tab:other_methods}. Our model was able to achieve an IoU of 34.67\% on the KITTI dataset \cite{geiger2012cvpr_kitti}. The other models have been trained on Semantic KITTI \cite{behley2019iccv_semantic_kitti} which is an annotated subset of the KITTI dataset. Their test IoU scores have been obtained on the scene completion task which ignores the semantic classes.
\section{Conclusion}
\label{sec:conc}
Significant progress has been achieved in the development of accurate and efficient monocular depth perception networks. However, these networks haven't been trained on complex unstructured road scenes, rendering them unsuitable for application in many developing countries. The proposed 3D occupancy convolution transformer network (OCTraN) paired with our dataset generation pipeline aims to mitigate this data distribution shift. These networks exhibit remarkable performance on a subset of the KITTI dataset and demonstrate tremendous potential in their capacity to generalize and construct dense 3D occupancy grids.

{\small
\bibliographystyle{ieee_fullname}
\bibliography{egbib}
}

\end{document}